\documentclass{article}

    \PassOptionsToPackage{numbers, compress}{natbib}

\usepackage{amsmath}
\usepackage{listings}
\usepackage{alltt}
\usepackage{xcolor}
\usepackage{tcolorbox}
\tcbuselibrary{skins,breakable}
\usepackage{paracol}
\definecolor{mygreen}{HTML}{6B8E2E}
\definecolor{lightgrayline}{HTML}{CFCFCF}

 \usepackage[preprint]{neurips_2026}

\usepackage[utf8]{inputenc} 
\usepackage[T1]{fontenc}    
\usepackage{hyperref}       
\usepackage{url}            
\usepackage{booktabs}       
\usepackage{amsfonts}       
\usepackage{nicefrac}       
\usepackage{microtype}      
\usepackage{xcolor}         
\usepackage{graphicx}
\usepackage{caption}
\usepackage{hyperref}
\usepackage{graphicx}
\usepackage{subcaption}
\usepackage{array}
\usepackage{tabularx}
\usepackage{makecell}
\usepackage{booktabs}
\usepackage{enumitem}
\usepackage{tikz}
\usepackage{algorithm}
\usepackage{needspace}
\usepackage{float}
\usepackage{algpseudocode}
\usepackage{wrapfig}
\usetikzlibrary{
    positioning,
    arrows.meta,
    decorations.pathreplacing,
    fit,
    backgrounds,
    shapes.geometric
}

\renewcommand{\cite}{\citep}

\usepackage{enumitem}
\setlist[itemize]{leftmargin=0.8cm}
\setlist[enumerate]{leftmargin=0.8cm}

\newcolumntype{C}[1]{>{\centering\arraybackslash}m{#1}}
\newcolumntype{L}[1]{>{\raggedright\arraybackslash}m{#1}}

\title{LLM Advertisement based on Neuron Auctions}

\author{%
  Peiran Yun\thanks{Equal contribution.} \\
  Tsinghua University \\
  Beijing, China \\
  \texttt{yunpr22@mails.tsinghua.edu.cn}
  \And
  Wenxin Xu\footnotemark[1] \\
  Tsinghua University \\
  Beijing, China \\
  \texttt{xuwx23@mails.tsinghua.edu.cn}
  \And
  Jiayuan Liu\footnotemark[1] \\
  Carnegie Mellon University \\
  Pittsburgh, PA, USA \\
  \texttt{jiayuan4@cs.cmu.edu}
  \And
  Yihang Zhang \\
  Tsinghua University \\
  Beijing, China \\
  \texttt{hemmaxacand@gmail.com}
  \And
  Liang Zeng \\
  Tsinghua University \\
  Beijing, China \\
  \texttt{zengliangcs@gmail.com}
  \And
  Lingkai Kong \\
  Harvard University \\
  Cambridge, MA, USA \\
  \texttt{lingkaikong@g.harvard.edu}
  \And
  Tonghan Wang\thanks{Correspondence to: Tonghan Wang <twang1@g.harvard.edu>} \\
  College of AI \\
  Tsinghua University \\
  Beijing, China \\
  \texttt{twang1@g.harvard.edu}
}

\begin{document}

\maketitle

\begin{abstract}
As Large Language Models (LLMs) transition into conversational agents, generative advertising emerges as a crucial monetization strategy. However, embedding advertisements within unstructured LLM outputs introduces a critical trilemma: balancing advertiser payoffs, platform revenue, and user experience. Existing methods, such as prompt injection or rigid position slots, disrupt semantic coherence and lack a parametric framework for independent control, rendering rigorous mechanism design intractable. 
To bridge this gap, we introduce \emph{Neuron Auctions}, a novel paradigm that shifts the auction object from the surface text space to the LLM's internal representations. 
Leveraging mechanistic interpretability, we identify brand-specific feed-forward network (FFN) neurons and demonstrate that competing brands activate within approximately orthogonal subspaces. This near-perfect independence allows us to define continuous, disentangled intervention budgets (specifically, neuron counts and amplification factors) as auctionable commodities. Building on this computational carrier, we design a continuous menu-based auction mechanism that naturally guarantees strategy-proofness and optimizes revenue for the platform. 
By explicitly incorporating a user utility penalty into the platform's optimization objective, our framework dynamically prices out overly aggressive interventions. Extensive experiments demonstrate that Neuron Auctions effectively preserve natural discourse quality while achieving an optimal alignment between commercial incentives and user satisfaction.
\end{abstract}

\section{Introduction}


The rapid adoption of Large Language Models (LLMs) is driving a fundamental paradigm shift in information retrieval, transitioning traditional search engines into conversational AI agents. This evolution naturally necessitates a corresponding shift in commercial monetization towards generative advertising~\citep{duetting2024mechanism, zhao2025llm}. Advertising in LLMs is embedded in an autoregressive generation process that jointly determines what information is presented and how it is surfaced. This introduces a new mechanism design challenge involving three stakeholders: advertisers, users, and the platform. Advertisers seek to maximize their Click-Through Rates (CTR) and Return on Investment (ROI) through highly visible recommendations; users value helpful, coherent, and high-quality responses; and platforms aim to maximize revenue while preserving user retention and satisfaction. A central question is reconciling these commercial and experiential objectives within the unstructured, autoregressive generation process of LLMs, mitigating the risk of models forsaking user welfare to heavily promote sponsored options \citep{wu2026ads}.


Existing approaches provide important first steps, but they do not fully address this challenge. Extensions of classical search advertising mechanisms, such as Generalized Second-Price (GSP) auctions~\citep{edelman2007internet, varian2007position}, are built on position-based allocation and naturally assume predefined advertising slots. While effective in traditional search settings, such slot-based designs are less suited to the fluid and context-dependent nature of LLM responses, where advertisements must be integrated natively into generated content. More recent work has explored prompt-based ad insertion methods~\citep{balseiro2026position,hajiaghayi2024ad}, offering a more flexible way to influence model outputs. However, these approaches typically do not provide a parametric mechanism for assigning fine-grained, independent recommendation strengths to competing brands, nor can they explicitly control user utility degradation. In addition, existing methods often inherit standard pricing rules such as GSP or VCG, leaving open the question of how to optimize platform revenue in the generative advertising setting.

We trace these limitations to a central difficulty in generative advertising: the choice of the auctionable object. In classical search advertising, the auctionable object is a position; in prompt-based approaches, it is an instruction or insertion opportunity. Both are defined in the surface text space, where control over generation is often coarse, entangled with response quality, and difficult to parameterize. In contrast, effective monetization in LLMs requires a computational carrier that is advertiser-specific, independently controllable, compatible with the model's native generation process, and, crucially, equipped with a measurable intervention budget that can be translated into allocation and pricing.

To this end, we shift the auction paradigm from surface-level text to the model's internal representation space. We introduce \textbf{\emph{Neuron Auctions}}, in which the auctionable commodities are selected feed-forward network (FFN) neurons. Building on recent advances in mechanistic interpretability~\citep{dai2022knowledge, turner2023activation}, we identify subsets of ``brand-specific neurons'' whose activations modulate the model's propensity to recommend particular advertisers. This representation-space formulation enables fine-grained control over recommendation strength through explicit intervention parameters, including the number of activated top-$k$ neurons and their amplification factor $\lambda$, while allowing the final response to be generated naturally by the LLM.

A key property enabling this design is separability. We show, both theoretically and empirically, that brand-specific representations tend to occupy approximately orthogonal subspaces. In high-dimensional activation spaces, this geometry implies that interventions for different advertisers interfere only weakly with one another. Consistent with this prediction, our experiments find little observable correlation or collision across brand interventions. This approximate independence makes FFN activations a natural computational carrier for generative advertising: \emph{advertiser-specific interventions can be parameterized, allocated, and priced without forcing advertisements into rigid slots or disrupting the fluidity of generation.}

This structure further provides the foundation for a mechanism design framework. The intervention budget---specified by the number of intervened neurons---acts as a controllable proxy for recommendation intensity. Empirically, increasing the intervention degree consistently strengthens the target brand's presence in generated responses, yielding a predictable relationship between allocation and advertising effect. This parametric relationship allows us to treat neurons as auctionable commodities and to design a continuous \emph{menu-based auction mechanism}. In our framework, the platform offers each advertiser a menu of prices and expected intervention levels, allowing advertisers to select the degree of recommendation strength that best matches their value. Prices are generated by a neural network that can be optimized to maximize an objective function.

We incorporate both platform revenue and a user utility into this objective, so that overly aggressive interventions are priced or discouraged when they degrade response quality. Our experiments show that platform revenue optimized by our method is significantly higher than
standard pricing rules such as GSP or VCG, and differs from approaches that approximate VCG-style payments through computationally intensive multi-fidelity optimization~\citep{clarke1971multipart,groves1973incentives,vickrey1961counterspeculation,liu2026incentive}. Together, the independence and priceability of neuron-level interventions allow \emph{Neuron} Auctions to address the generative advertising trilemma: improving advertiser visibility, preserving users' conversational experience, and optimizing platform revenue within a unified framework.

\section{Related Work}
\label{sec:related_work}

\paragraph{Mechanism Design for Generative Advertising}
Traditional search advertising has been heavily influenced by the Generalized Second-Price (GSP) auction \cite{edelman2007internet,varian2007position} and the truthful Vickrey-Clarke-Groves (VCG) mechanism \cite{clarke1971multipart,groves1973incentives,vickrey1961counterspeculation}. However, integrating these classic mechanisms into generative AI requires a paradigm shift: traditional position-based models are incompatible with the unstructured, autoregressive nature of LLMs, and calculating exact VCG payments demands computationally prohibitive counterfactual evaluations. To directly address this computational bottleneck, \citet{liu2026incentive} coupled VCG incentives with multi-fidelity optimization to maximize expected social welfare under strict generative budgets while establishing formal guarantees for approximate strategy-proofness.
Parallel to such computational optimizations, recent literature has actively redefined the auction object itself. \citet{duetting2024mechanism} introduced the foundational token auction model via single-dimensional bids, where advertisers bid token-by-token to insert their preferred words. 
Subsequently, ad auctions have been customized for Retrieval-Augmented Generation (RAG) systems \citep{hajiaghayi2024ad}, and position auctions in continuous AI-generated content have been formalized by explicitly modeling substitution effects through multinomial logit and cascade user behaviors \citep{balseiro2026position}. To further mitigate latency and privacy issues, other frameworks either treat the LLM output distribution as the auction object via reward-preference optimization \citep{zhao2025llm}, or completely decouple ad insertion from generation by utilizing semantic ``genres'' as a proxy for VCG auctions \citep{xu2026ad}. 

Yet, introducing competing commercial objectives poses severe alignment risks, as models may forsake user welfare to heavily promote sponsored options \citep{wu2026ads}. More importantly, whether utilizing decoupled frameworks or token-level bidding, relying on surface-level text insertions or external prompt manipulations fundamentally disrupts the natural discourse and semantic coherence of the generated response. Unlike these methods, our approach shifts the auction paradigm directly to the internal representation space of the LLM, maintaining strategy-proofness while explicitly bounding user utility degradation through native neuron interventions.

\paragraph{Influencing LLMs and Internal Intervention}

To achieve fine-grained, independent control over brand recommendations without prompt manipulation, we must consider how to best influence modern LLMs. During post-training, parameter-efficient fine-tuning (PEFT, e.g., LoRA) inserts small modules that achieve strong effects \citep{hu2022lora,pfeiffer2021adapterfusion}. However, deliberate curation of duplicated data, full pretraining, RLHF, or DPO hard-wire behaviors \citep{ouyang2022training,rafailov2023direct}, which are impractical for rapid, per-query, multi-stakeholder advertising. At inference, system prompts can be brittle, and plug-and-play decoding steers generation via auxiliary scores at a noticeable latency cost \citep{dathathri2020plug,krause2021gedi}. 

Therefore, our intervention mechanism leverages recent breakthroughs in mechanistic interpretability. Previous work has demonstrated that feed-forward network (FFN) layers function as key-value memories \citep{dai2022knowledge}. Recent techniques like Activation Addition (ActAdd) and Sparse Autoencoders (SAEs) edit interpretable features on-the-fly, offering continuous control knobs on influence strength without weight updates \citep{turner2023activation,obrien2024sae,koriagin2025teach}. Our method is highly synergistic with contrastive activation addition \citep{panickssery2023steering}, utilizing cloze-prompt differences to isolate brand neurons. By extending these steering techniques into an economic setting, we demonstrate that FFN activations can be disentangled and auctioned as independent commodities.

\section{Neuron Auction Framework}

We consider an LLM platform that mediates $n$ advertisers, where advertiser $a$ is associated with a brand $b_a$. Our goal is to define an auctionable object that can modulate the model's tendency to recommend a sponsored brand while preserving the natural autoregressive generation process. To this end, rather than allocating surface-level positions or appending advertising prompts, we intervene on a small set of brand-related feed-forward network (FFN) neurons in the LLM.

For each brand $b$, we first identify a ranked set of brand-specific neurons using gradient-based attribution. Let $\Omega_b(k)$ denote the top-$k$ neurons most strongly associated with brand $b$. An intervention option for advertiser $a$ is parameterized by the number of intervened neurons, $k_a$. During response generation, the input prompt remains unchanged. Instead, for each neuron $i$ in layer $l$, we modify its FFN activation as
\[
\tilde{h}_{i}^{(l)}
=
\begin{cases}
\lambda h_{i}^{(l)}, & i \in \Omega_{b_a}(k_a),\\
h_{i}^{(l)}, & i \notin \Omega_{b_a}(k_a).
\end{cases}
\]
Here, $h_i^{(l)}$ is the original intermediate activation, $\tilde{h}_i^{(l)}$ is the intervened activation, and $\lambda$ is an amplification coefficient. Thus, the allocation variable \(k_a\) directly controls the intervention degree.

This representation-space formulation has three properties that make it suitable for auction design. First, the intervention is advertiser-specific: each bidder is associated with a brand-dependent ranked neuron set. As we show later, our construction of brand-specific neuron sets is also approximately separable across advertisers, enabling largely independent control of competing brand interventions. Second, the allocation is measurable and ordered: larger values of \(k_a\) correspond to stronger intervention levels, allowing the platform to construct menus over recommendation intensity. Third, because the response is still generated autoregressively by the LLM, the intervention avoids rigid ad slots or direct prompt manipulation and better preserves the fluency and coherence of the generated answer.

\subsection{Intervention Method}
\label{sec:intervention}

\noindent\textbf{Offline attribution scoring.}
The first step is to identify neurons that are selectively associated with each brand. For every brand $b$, we calculate an attribution score for each FFN neuron, where a larger score indicates that the neuron more strongly promotes brand $b$. The resulting ranked neuron list defines the candidate intervention set from which the platform later constructs auction options. The key challenge is specificity: a useful neuron should be highly responsive to the target brand while being weakly responsive to competing brands.


\begin{figure}[t]
    \centering
    \includegraphics[width=1.0\linewidth]{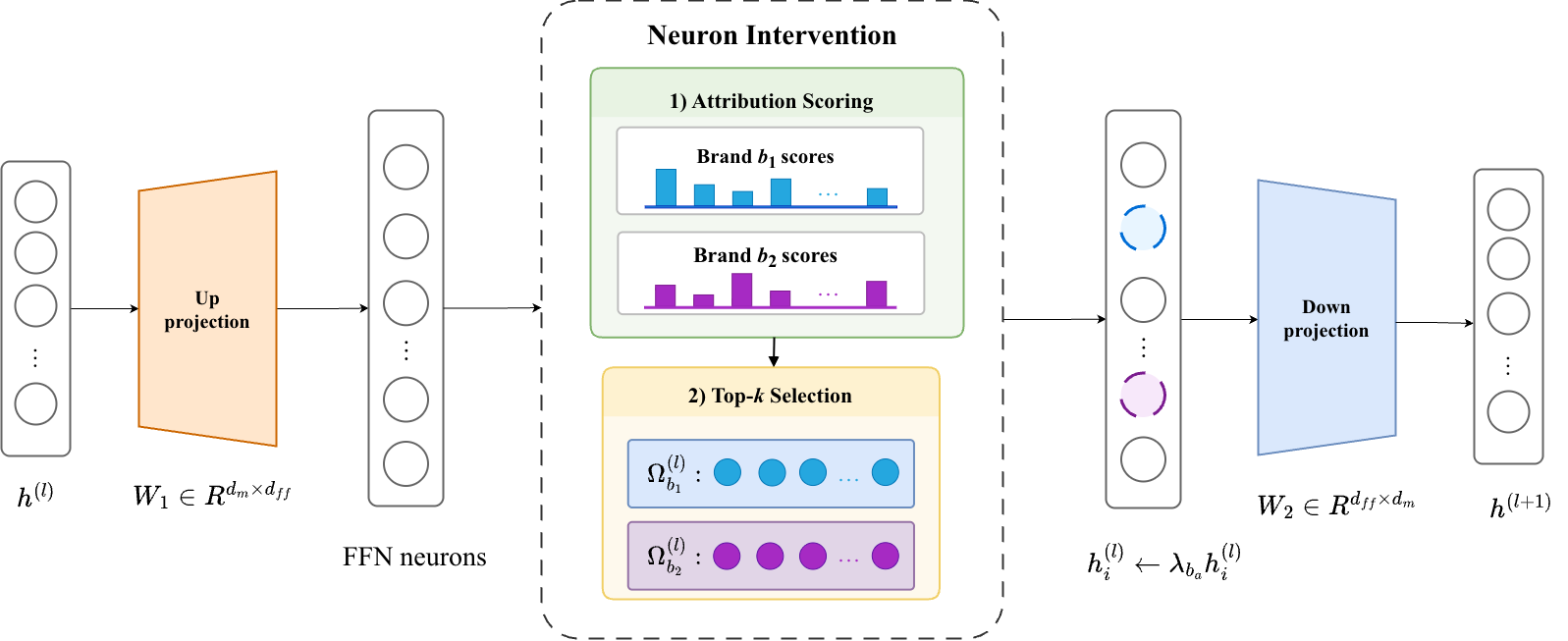}
    \caption{Overview of the proposed neuron intervention pipeline in the FFN at layer $l$. $\Omega_{b_a}^{(l)}$ stands for the Top-$k$ Neurons chosen to be amplified for bidder $b_a$ in Layer $l$, and $\lambda_{b_a}$ is the intervention coefficient. }
    \label{fig:intervetion_pipeline}
    
\end{figure}

We adopt a cloze-style attribution strategy to calculate brand-level scores. For each brand $b$, we construct $M$ semantically similar but lexically diverse cloze prompts, denoted by
\(
\mathcal{P}_b=\{p^{(b)}_1,\dots,p^{(b)}_M\}.
\)
All prompts are designed to elicit the same brand concept. Using a single prompt can introduce substantial contextual noise, making attribution scores sensitive to superficial prompt variations. Averaging over multiple synonymous cloze prompts reduces this variance and emphasizes features consistently associated with the brand itself. Examples of cloze prompts are provided in Appendix~\ref{app:clozes}.

However, neuron intervention methods~\citep{dai2022knowledge} typically define attribution with respect to a single target token. This is insufficient for brand names that are tokenized into multiple sub-tokens. For example, Qwen3 tokenizes ``Hyatt'' as \texttt{['Hy', 'att']}. We therefore use the log-probability of the full brand token sequence as the attribution target.

Let $y^{(b)}=(y_1,\dots,y_T)$ be the tokenized brand and $p'_t=p\text{ || } y_{<t}$ be the prefix for predicting token $y_t$.
For cloze prompt $p$, the attribution score of neuron $(l,i)$ is defined as
\begin{equation}
A_i^{(l)}(b;p)
=
\sum_{t=1}^{T}
h_i^{(l)}(p'_t)
\int_0^1
\frac{
\partial
\log P\!\left(y_t \mid p'_t; \alpha h_i^{(l)}(p'_t)\right)
}{
\partial \left(\alpha h_i^{(l)}(p'_t)\right)
}
\, d\alpha .
\label{eq:logp-attribution}
\end{equation}
Here, $h_i^{(l)}(p'_t)$ is the FFN intermediate activation at the last token position of $p'_t$.
The log-probability form decomposes the attribution of a multi-token brand into additive token-wise terms. It also avoids the numerical instability caused by multiplying several token probabilities.

In addition, to further identify neurons that are truly brand-specific, we construct for each brand $b$ a set of competing brands, denoted by $\mathcal{N}(b)=\{b_1^-,\dots,b_K^-\}$, and adopt a contrastive strategy. Specifically, we first average the attribution of the target brand across multiple cloze prompts:

$$
S_i^{(l)}(b) = \frac{1}{|\mathcal{P}_b|} \sum_{p\in \mathcal{P}_b} A_i^{(l)}(b;p),
$$

and then subtract the average attribution of the same neuron to competing brands:

$$
\tilde{S}_i^{(l)}(b) = S_i^{(l)}(b) - \frac{1}{|\mathcal{N}(b)|} \sum_{b^- \in \mathcal{N}(b)} S_i^{(l)}(b^-) .
$$

The larger this score is, the more likely neuron $(l,i)$ is to be a ``true positive neuron'' for the target brand, meaning that it tends to promote the generation of the target brand rather than generally supporting the outputs of brands in the same category.

Since the exact computation of the integral is expensive, we approximate it using a Riemann sum with $m$ discrete steps. The attribution score used in practice is therefore written as
$$
A_i^{(l)}(b;p)
\approx
\sum_{t=1}^{T}
\frac{h_i^{(l)}(p'_t)}{m}
\sum_{r=1}^{m}
\left.
\frac{
\partial
\log P\!\left(y_t \mid p'_t; \alpha h_i^{(l)}(p'_t)\right)
}{
\partial \left(\alpha h_i^{(l)}(p'_t)\right)
}
\right|_{\alpha=r/m}.
$$
In our experiments we set $m=20$, achieving a good balance between computational cost and attribution stability. Through the above procedure, for any brand, we are able to reliably identify its corresponding ``true positive neurons''. 

\noindent\textbf{Online Neuron Selection and Intervention}. \ \ After obtaining neuron attribution scores for each brand, we further perform a second-stage filtering procedure based on \emph{brand specificity}. Suppose that there are $N$ bidders participating in one auction round, denoted by $\{b_1,\dots,b_N\}$. For a candidate neuron $i$ associated with brand $b_j$, we define its online ranking score as
$$
\mathrm{Attr}_i^{(l)}(b_j)
=
\mathrm{\tilde{S}}_i^{(l)}(b_j)
-
\frac{\sum_{b' \neq b_j}\left|\mathrm{\tilde{S}}_i^{(l)}(b')\right|}{N-1}.
$$

Specifically, by subtracting the absolute attribution scores of all non-target brands, we penalize neurons that are also highly associated with competing brands, regardless of whether their effects are promotional or inhibitory. Based on this criterion, we rank neurons separately for each brand and select the top-$k$ neurons to form the final intervention set. During generation, we only amplify the selected neurons at the last token of the current decoding step, which is consistent with the autoregressive decoding process of Transformer decoders. Let $\Omega_b$ denote the target neuron set associated with brand $b$, and let $\lambda_b > 1$ denote the amplification factor. We then intervene on the FFN intermediate activations of the last token as follows:
$$
h_i^{(l)} \leftarrow \lambda_b h_i^{(l)}, \quad (l,i)\in \Omega_b.
$$

However, when multiple bidders simultaneously inject their brand-specific neurons, independently amplifying all selected neurons often causes the hidden-state norm to grow rapidly, which can further lead to response collapse. To address this issue, we design a \emph{unified-hook} mechanism:  we first apply amplification to the target neurons of all brands, and then perform a unified norm rescaling on the modified last-token representation so that its overall $L_2$ norm is restored to the pre-intervention level. 
$$
h^{(l)}_{new} = \frac{||h^{(l)}_{old}||_2}{||h^{(l)}_{intervened}||_2} \cdot h^{(l)}_{intervened} ,
$$

where $h^{(l)}_{\mathrm{old}}$ denotes the original FFN intermediate activation vector at the last token position in layer $l$, $h^{(l)}_{\mathrm{intervened}}$ denotes the vector after applying all selected neuron amplifications, and $h^{(l)}_{\mathrm{new}}$ denotes the final rescaled vector fed back into the model.
This rescaling controls the overall magnitude of the FFN intermediate representation after intervention. Without this normalization, simultaneously amplifying neurons for multiple brands can substantially increase the activation norm, pushing the representation away from the typical activation distribution and causing unstable or repetitive decoding. 

In this way, we increase the probability that the target brand is mentioned and recommended in natural responses, while preserving generation stability.

\subsection{Neuron Auctions}
\label{sec:virtual}

We now define the auction mechanism built on top of neuron-level interventions. The auctionable object is the intervention degree $k$: the number of brand-specific neurons whose activations are amplified during generation. As validated in Section~\ref{sec:intervention_exp}, larger values of $k$ lead to stronger recommendation effects for the target brand. Thus, $k$ provides an ordered and quantifiable proxy for advertising intensity. Building on this intervention parameter, we adopt a menu-based auction representation~\citep{wang2024gemnet} allowing the platform to price different levels of recommendation strength directly in the model's representation space.

For each advertiser $i$, the platform offers a menu
\begin{align}
\mathcal{M}_i(v_{-i};\theta)
=
\left\{
\left(k_i^{(j)}, p_i^{(j)}(v_{-i};\theta)\right)
\right\}_{j=1}^{J},    
\end{align}
where each option consists of an allocation level $k_i^{(j)}$ and a corresponding price $p_i^{(j)}$. The allocation $k_i^{(j)}$ specifies how many brand-specific neurons of advertiser $i$ will be amplified, thereby determining the intervention degree and the induced advertising effect. We denote the resulting click-through rate by $C(k_i^{(j)})$. Given the menu, advertiser $i$ selects the option that maximizes its utility:
\begin{align}
    u_i^{(j)}(v_{-i})
=
v_i \cdot C(k_i^{(j)})
-
p_i^{(j)}(v_{-i};\theta),\ \ \ \ \ \ \ \ \ \ \ \ 
j_i^*
=
\arg\max_j u_i^{(j)}(v_{-i}).
\end{align}
The platform then allocates intervention degree $k_i^{(j_i^*)}$ and charges price $p_i^{(j_i^*)}(v_{-i};\theta)$. The allocation levels $\{k_i^{(j)}\}_{j=1}^J$ are fixed menu options, while the prices are generated by a neural network parameterized by $\theta$. The input to this neural network includes the bidder index and the bids submitted by all other advertisers, $v_{-i}$. We also include an option with zero allocation and zero price, ensuring that advertiser utility is not smaller than zero (individual rationality). The price of this option is not trainable.

\emph{User utility.} The selected intervention degrees jointly determine the final ad-integrated response generated by the LLM. Let
\(
q\left(\{k_i^{(j_i^*)}\}_{i=1}^n\right)
\)
denote the estimated user-side quality of this response. This term captures whether the generated answer remains helpful, coherent, and responsive to the user's query after neuron interventions. Both the click-through function $C(\cdot)$ and the user-quality function $q(\cdot)$ are estimated using the LLM-based scoring procedure described in Appendix~\ref{appx:llm-judge}, which combines dimension-specific evaluation prompts, evidence-before-score reasoning, few-shot human-calibrated exemplars, and deterministic decoding to provide stable and interpretable proxy measurements.

The platform optimizes prices to balance commercial revenue and user experience. We define the platform objective as
\begin{align}
\mathcal{J}(\theta)
=
\sum_i \sum_j z_i^{(j)} p_i^{(j)}(v_{-i};\theta)
+
w_{\mathtt{user}}
q\left(\{k_i^{(j_i^*)}\}_{i=1}^n\right),
\label{eq:objective}
\end{align}
where
\(
z_i^{(j)}
=
\mathrm{SoftMax}_j
\left(
u_i^{(1)}(v_{-i}),\ldots,u_i^{(J)}(v_{-i})
\right)
\)
is a differentiable approximation of advertiser $i$'s menu choice, and $w_{\mathtt{user}}$ controls the relative importance of user experience. The first term is thus a differentiable approximation of the platform revenue. Training is implemented by updating the learnable parameters $\theta$ to minimize the loss $-\mathcal{J}(\theta)$.

By explicitly incorporating the user-quality term into the platform objective, the mechanism learns to price intervention levels according not only to advertisers' willingness to pay, but also to the degradation they impose on the generated response. As a result, the platform can discourage overly intrusive advertising while still allocating recommendation strength to advertisers who value it most.

\section{Experiments}
\label{experiments}
\subsection{Intervention Experiments}
\label{sec:intervention_exp}

To further validate the neuron intervention method proposed in
Section~\ref{sec:intervention}, we conduct intervention experiments
on \textbf{100 two-bidder combinations} across three
large language models, Qwen3-4B~\cite{yang2025qwen3}, DeepSeek-R1-8B~\cite{guo2025deepseek}, and Llama-3-8B~\cite{grattafiori2024llama},
as well as on \textbf{20 three-bidder combinations} on Qwen3-4B.
In all experiments, both attribution and intervention are performed
in the FFN space.

For each two-bidder combination $(b_1,b_2)$, we manually design $p=3$ prompts whose semantic contexts are naturally compatible with both brands. We first compute the contrastive attribution scores for the two advertisers separately and rank neurons accordingly, then select the top-$k$ neurons for each advertiser and perform joint intervention during generation using the unified-hook mechanism. In the two-bidder setting, we sweep the intervention budgets over $k_1, k_2 \in \{0,100,200,\dots,800\}.$ For each combination of three-bidder $(b_1,b_2,b_3)$ in Qwen3-4B, we follow the same procedure. Specifically, we design prompts that are semantically compatible with all three advertisers, compute attribution scores for each advertiser independently, and jointly intervene on their selected neurons during generation. To control the overall generation cost, in the three-bidder setting we use a slightly smaller intervention range: $k_1, k_2, k_3 \in \{0,100,200,\dots,600\}.$

We initially evaluate advertising effectiveness using the Q4 score defined in Appendix~\ref{app:llm scoring rule}, which serves as a principled and fine-grained proxy for user CTR. However, applying the LLM evaluator to all experiments is computationally expensive and may introduce additional evaluation noise. We therefore examine whether hit count can also serve as a lightweight metric for measuring advertising effectiveness. Empirically, we find that the heatmaps produced by hit counts exhibit patterns highly similar to those produced by the LLM evaluator, as shown in Figure~\ref{fig:q4_two_heatmaps}. Moreover, the hit-count heatmaps are often smoother and contain fewer noisy fluctuations than the evaluator-score heatmaps. Based on these observations, we adopt hit count as the evaluation metric in all subsequent experiments.

\begin{figure}[htbp]
    \centering

    \includegraphics[width=1\linewidth]{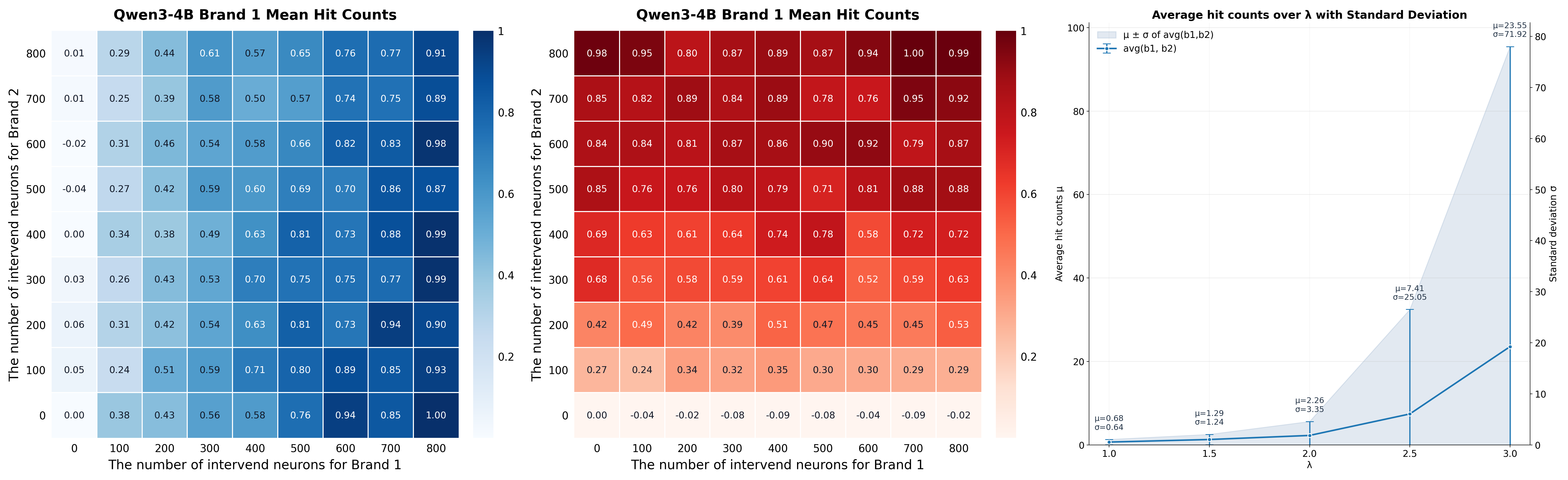}
    \caption{
    Heatmaps and ablation results of brand hit counts under joint neuron intervention for two advertisers on Qwen3-4B.
    The left and middle panels show the normalized hit-count changes for bidder 1 and bidder 2, respectively, with the x- and y-axes indicating the numbers of top-$k$ neurons selected for the two bidders.
    The right panel reports the ablation study over the intervention coefficient $\lambda$, where $\mu$ and $\sigma$ denote the mean hit count and standard deviation.
    All heatmap values are averaged over two-bidder combinations, normalized per bidder, shifted by the $(0,0)$ baseline, and rescaled so that the maximum positive change equals 1.
    }
    \label{fig:qwen_lambda}
\end{figure}

\begin{figure}[t]
    \centering

    \begin{subfigure}{0.32\linewidth}
        \centering
        \includegraphics[width=\linewidth]{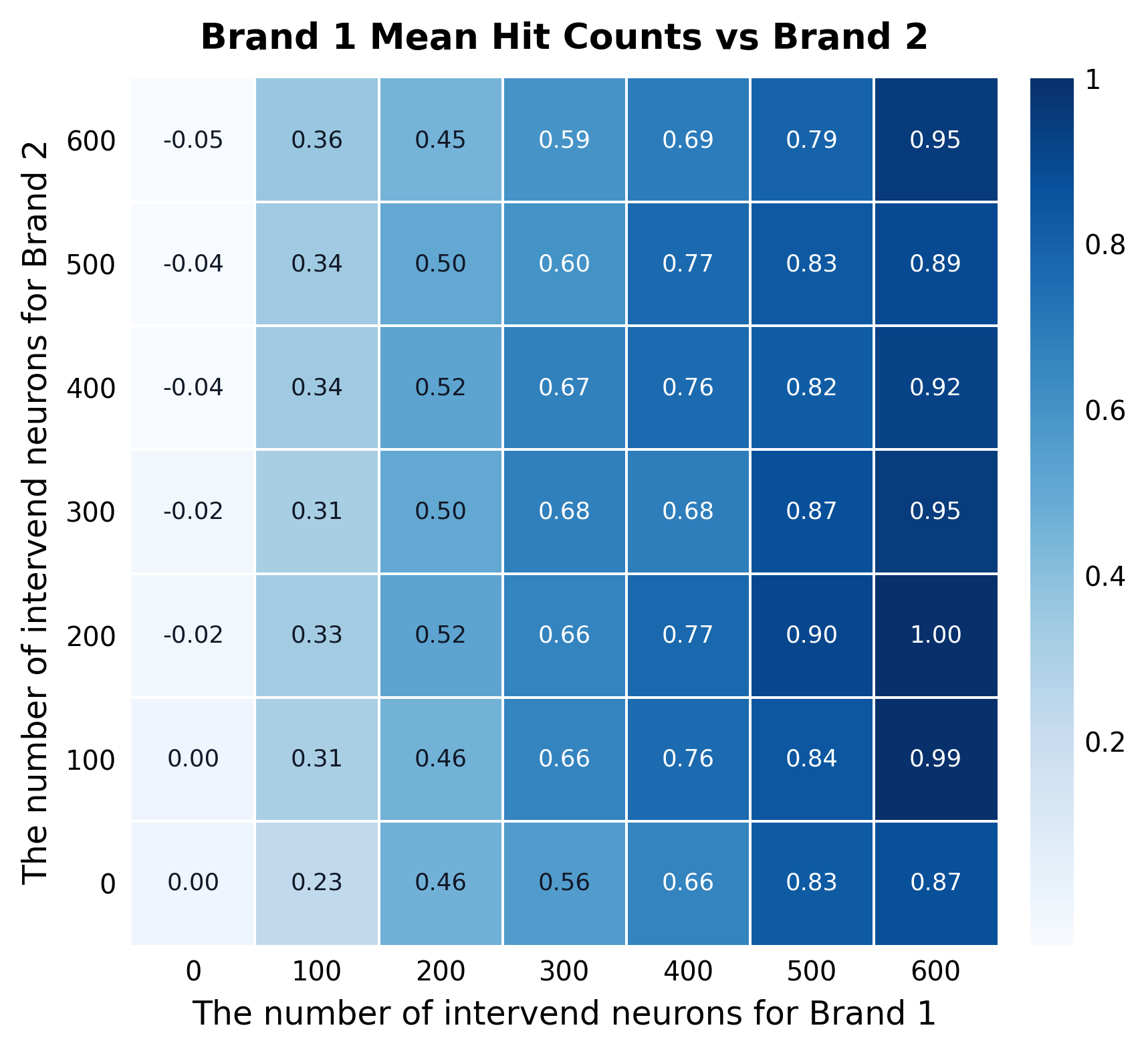}
        \caption{Brand 1 hit counts of $(k_1,k_2)$.}
        \label{fig:qwen_three_1vs2_b1}
    \end{subfigure}
    \hfill
    \begin{subfigure}{0.32\linewidth}
        \centering
        \includegraphics[width=\linewidth]{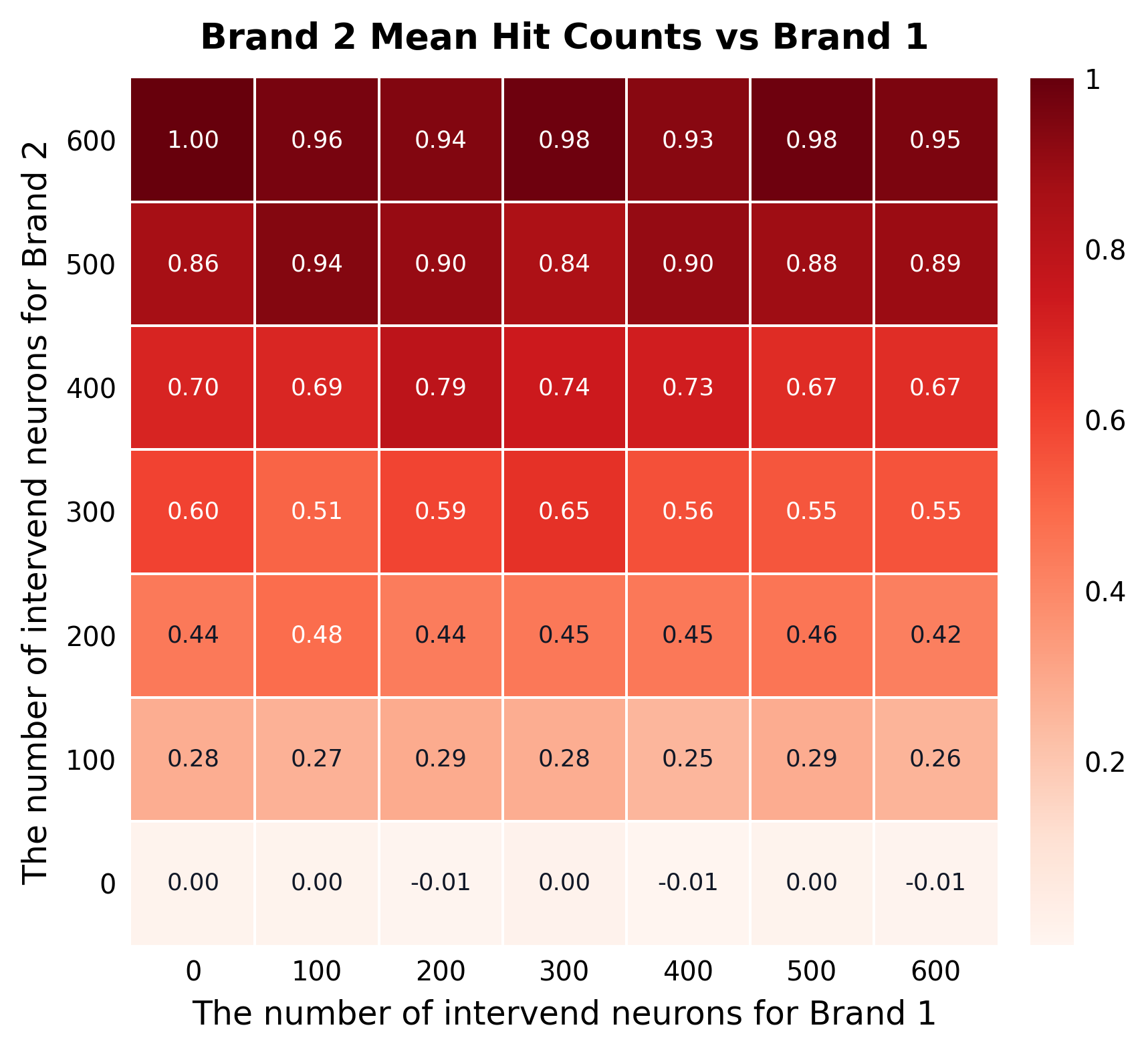}
        \caption{Brand 2 hit counts of $(k_1,k_2)$.}
        \label{fig:qwen_three_1vs2_b2}
    \end{subfigure}
    \hfill
    \begin{subfigure}{0.32\linewidth}
        \centering
        \includegraphics[width=\linewidth]{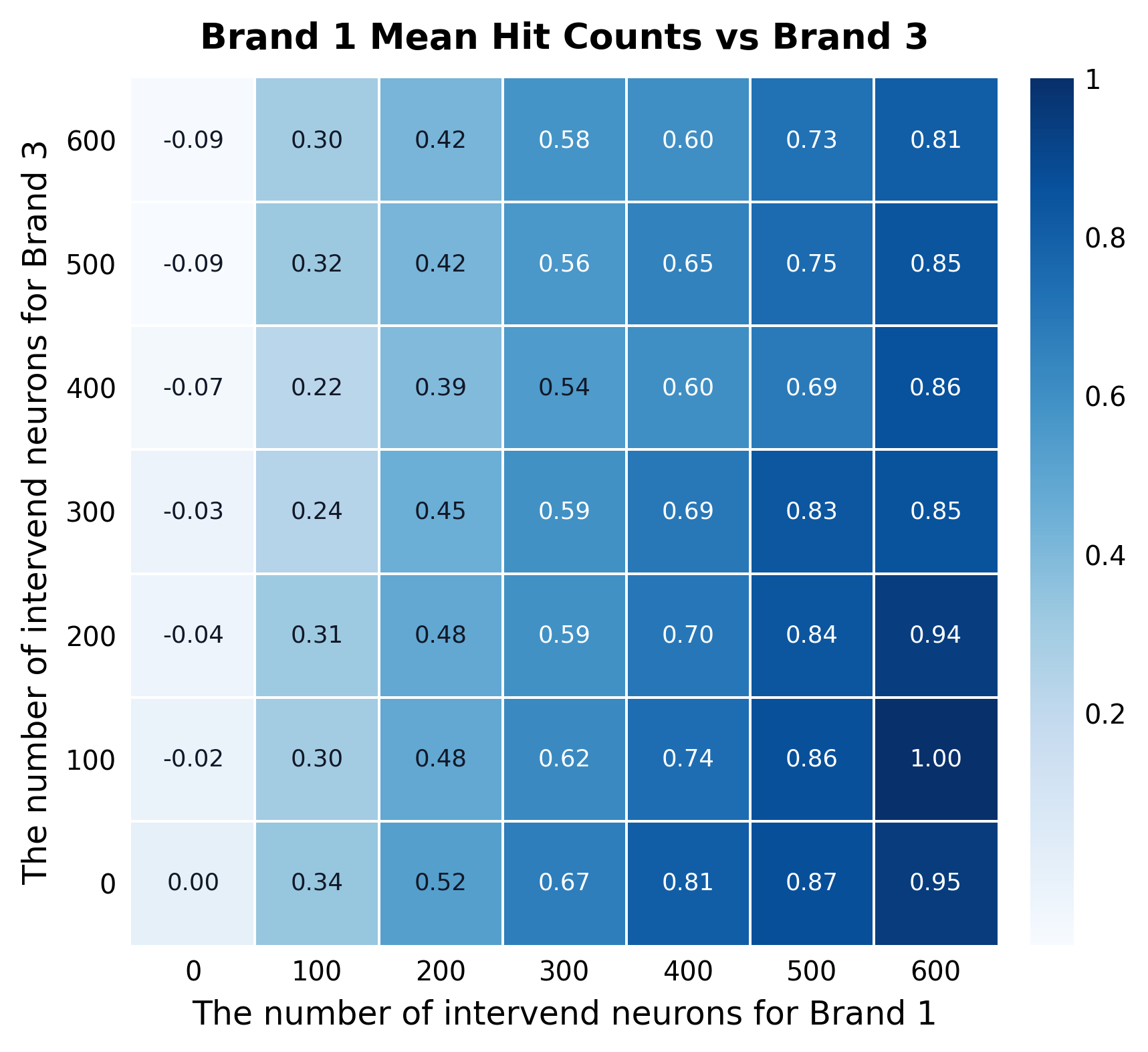}
        \caption{Brand 1 hit counts of $(k_1,k_3)$.}
        \label{fig:qwen_three_1vs3_b1}
    \end{subfigure}

    \vspace{0.5em}

    \begin{subfigure}{0.32\linewidth}
        \centering
        \includegraphics[width=\linewidth]{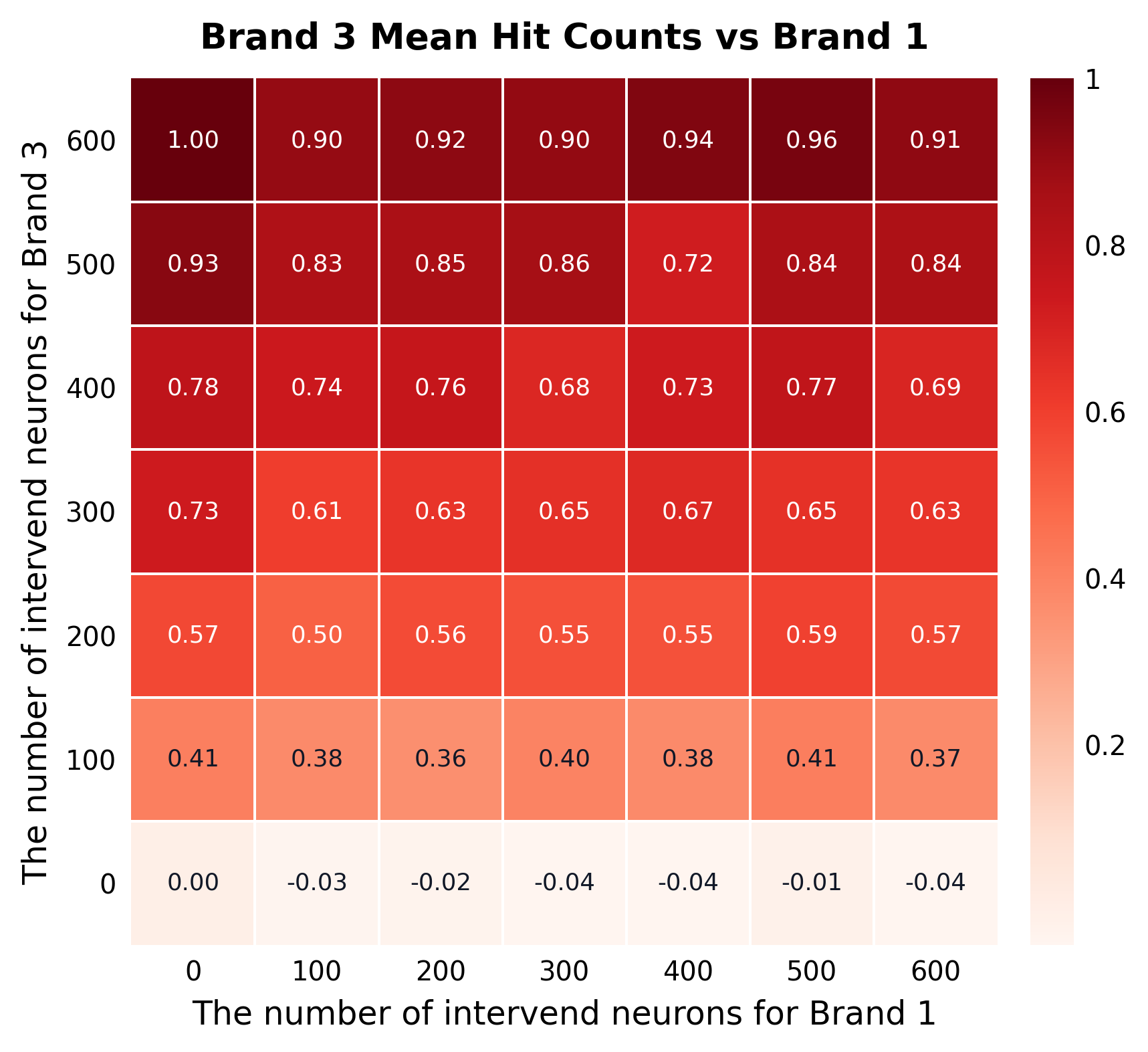}
        \caption{Bidder 3 hit counts of $(k_1,k_3)$.}
        \label{fig:qwen_three_1vs3_b3}
    \end{subfigure}
    \hfill
    \begin{subfigure}{0.32\linewidth}
        \centering
        \includegraphics[width=\linewidth]{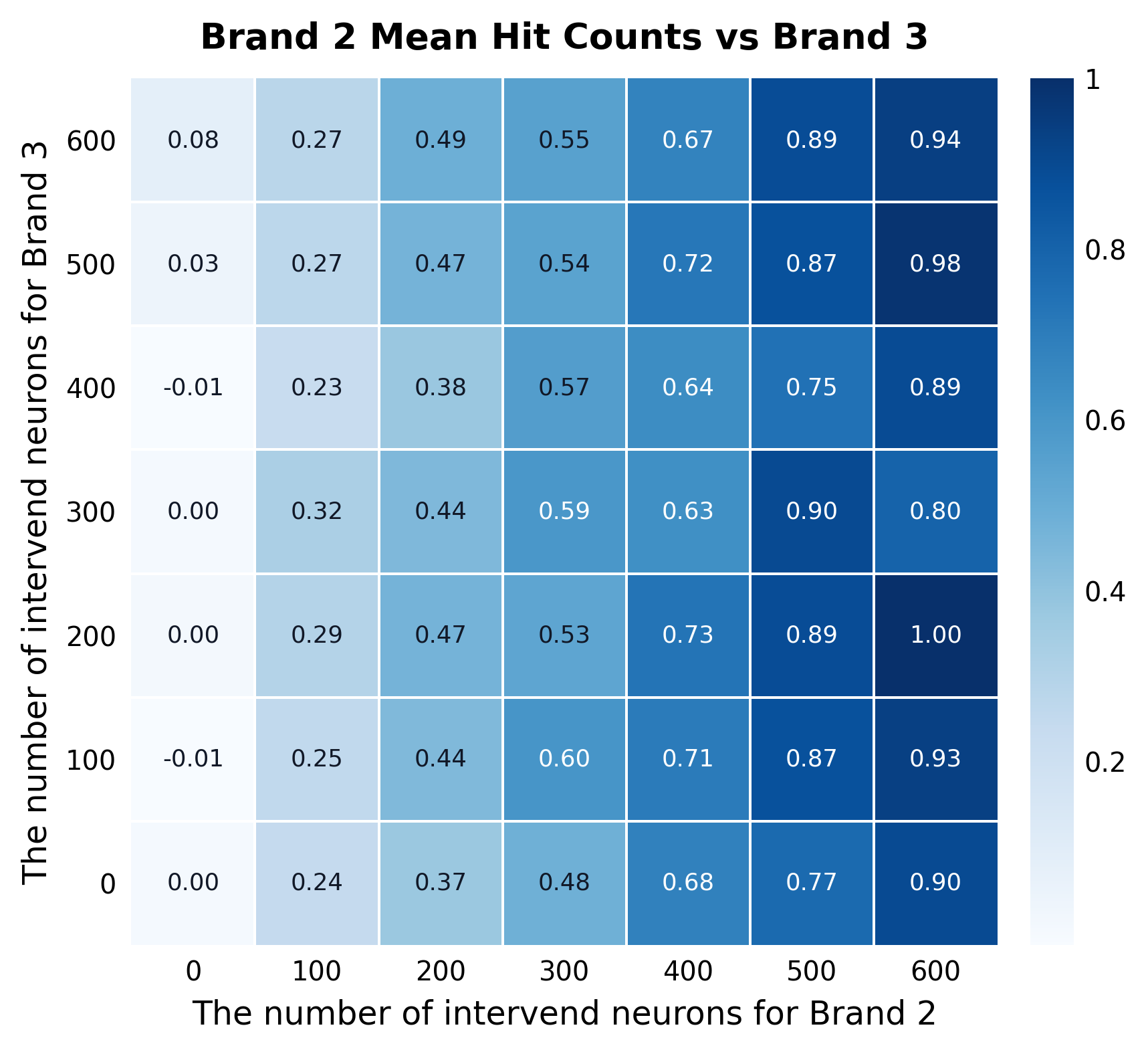}
        \caption{Brand 2 hit count of $(k_2,k_3)$.}
        \label{fig:qwen_three_2vs3_b2}
    \end{subfigure}
    \hfill
    \begin{subfigure}{0.32\linewidth}
        \centering
        \includegraphics[width=\linewidth]{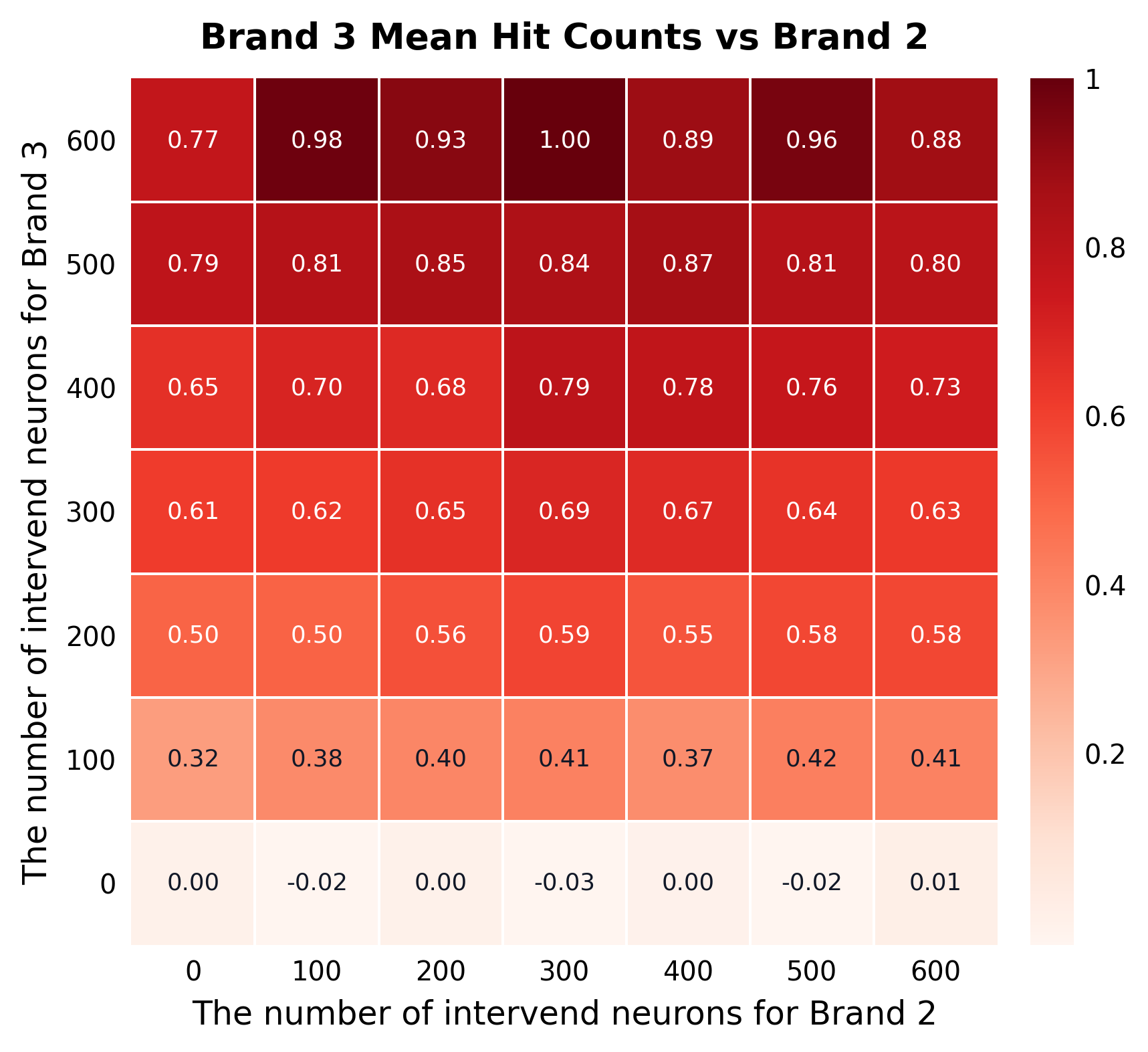}
        \caption{Brand 3 hit count of $(k_2,k_3)$.}
        \label{fig:qwen_three_2vs3_b3}
    \end{subfigure}

    \caption{Heatmaps of brand hit counts under joint neuron intervention for three bidders on Qwen3-4B. 
    Each subfigure shows the normalized hit counts of one bidder on a two-bidder intervention plane.}
    \label{fig:qwen_three_bidder_heatmaps}
\end{figure}

The results of the two-bidder and three-bidder experiments on Qwen3-4B are summarized in Figure~\ref{fig:qwen_lambda} and Figure~\ref{fig:qwen_three_bidder_heatmaps}. In both settings, we observe that increasing the intervention budget of a target advertiser consistently strengthens its presence in the generated responses, while cross-advertiser interactions remain secondary. For the two-bidder setting, the corresponding heatmaps for DeepSeek-R1-8B and Llama-3-8B also exhibit similar results, and their heatmaps are provided in Appendix~\ref{app:full_results}.


As for the intervention coefficient $\lambda$, we conduct an ablation study on
30 two-bidder combinations using Qwen3-4B, as shown in
Figure~\ref{fig:qwen_lambda}. We find that when $\lambda \leq 1.5$,
the intervention effect is relatively weak and the target brands are not
sufficiently promoted in the generated responses. In contrast, when
$\lambda=3.0$, the model frequently collapses into repetitive outputs of the
target brand names, which leads to sharp increases in both the average hit count
and its variance. Therefore, considering both recommendation effectiveness and
generation stability, we set a shared intervention coefficient
$\lambda_{b_a}=2.0$ for each bidder $a$ in all experiments.

The heatmaps in Figure~\ref{fig:qwen_lambda} and \ref{fig:qwen_three_bidder_heatmaps} further demonstrate that our intervention method is well aligned with the advertising task: as the number of intervened neurons increases, the target brand is mentioned substantially more often in the model's responses. Moreover, although joint intervention inevitably induces cross-brand interaction, the dominant factor for each brand remains its own allocated neuron budget. This indicates that the proposed attribution-based neuron selection preserves a substantial degree of brand specificity even in the multi-bidder setting, thereby providing a theoretical foundation for our earlier formulation that treats neurons as auction items.

\subsection{Menu Training Experiments}

Recall that the reward function designed earlier consists of \(n+1\) components: \(n\) components correspond to the revenue derived from each advertiser, and one component accounts for user experience. In the subsequent experiments, we assume that each bidder's valuation for CTR is uniformly and randomly sampled from the interval $[0,1]$.

\begin{figure}[htbp]
    \centering
    \includegraphics[width=0.95\textwidth]{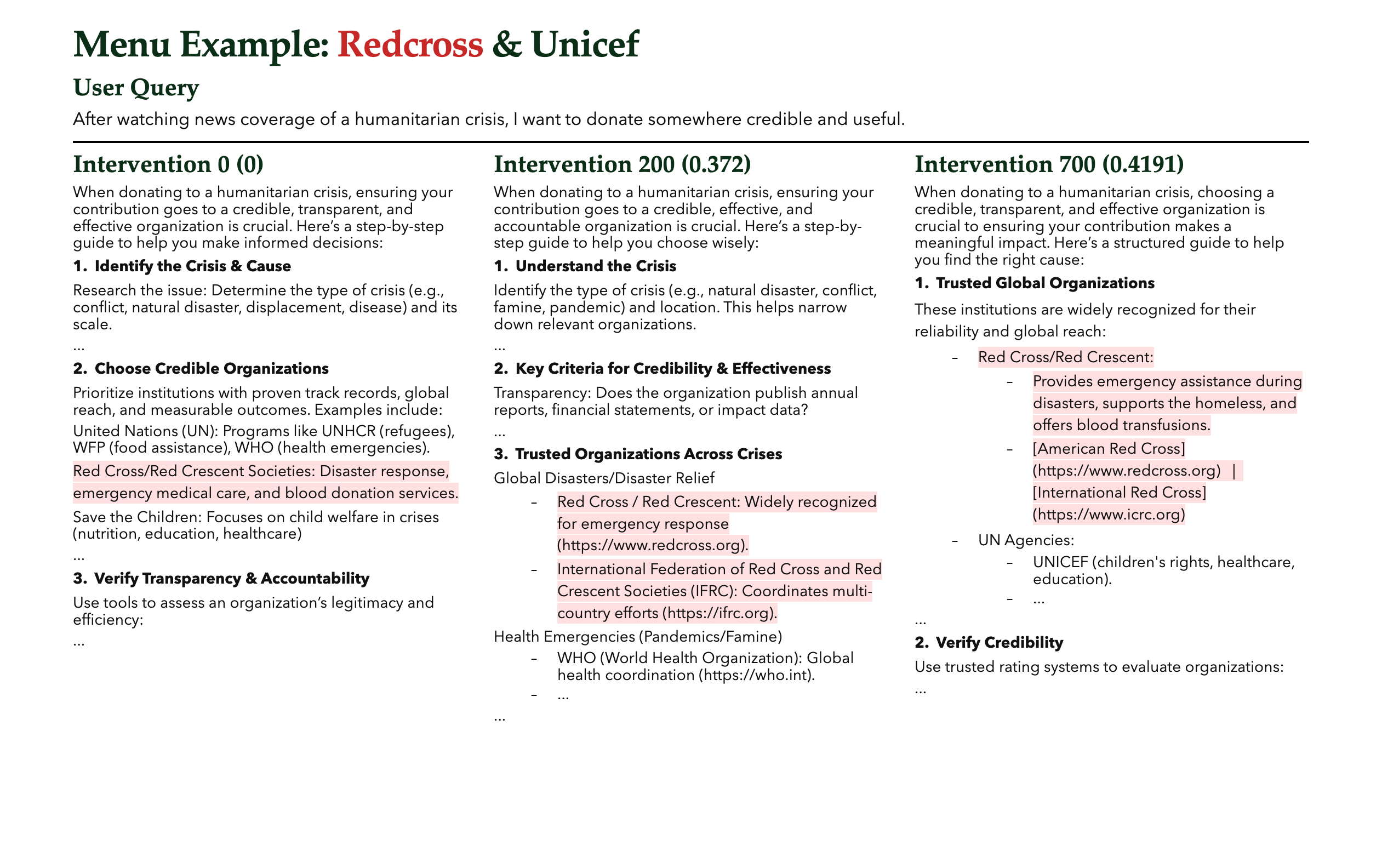}
    \caption{Example of Different Menu Options. Here, we fix the intervention strength of Unicef and focus on Redcross. Among all available menu options, the three options shown in the figure are those that can be selected under certain values. We highlight the advertisement-related parts. }
    \label{fig:menu_example}
\end{figure}

Figure~\ref{fig:menu_example} illustrates an example of the menu, where each option can be traversed under different bid values from the advertisers. The first option with a price of 0 only briefly mentions Redcross. By contrast, the second option with a price of 0.372 provides more detailed information including the official website link. Further, the third option with a price of 0.4191 places the advertising content much earlier in the response compared with the second one.

\begin{figure}[t]
    \centering
    \includegraphics[width=\textwidth]{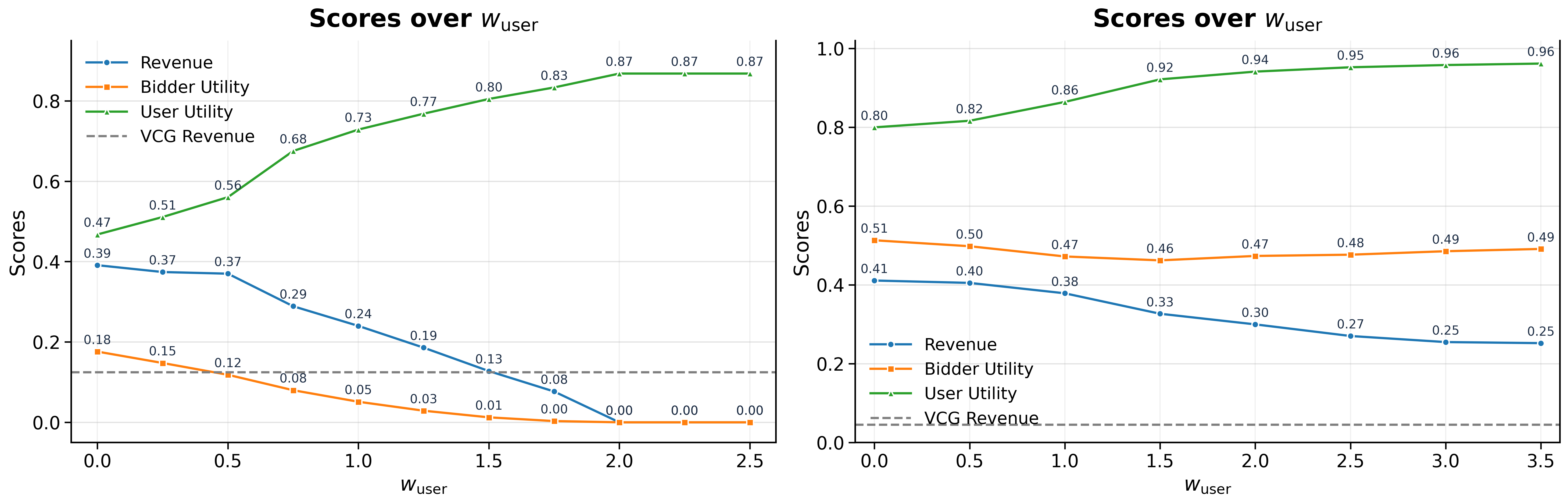}
    \caption{Scores over $w_{\text{user}}$. Here, Revenue denotes the total payoff obtained by the platform from the bidders, Bidder Utility denotes the sum of all bidders' rewards after subtracting their prices, and User Utility denotes the user's satisfaction with the generated response. VCG Revenue denotes the revenue obtained under the VCG mechanism when $w_{\mathrm{user}}=0$.}
    \label{fig:bidders_results}
\end{figure}

 To demonstrate the impact of $w_{\text{user}}$ on the overall model, we conducted comparisons across different settings both on cases of 2-bidder and 3-bidder, as shown in Figure ~\ref{fig:bidders_results}. The results show that $w_{\mathrm{user}}$ serves as an effective control parameter for regulating the trade-off between monetization and user experience. In both cases, increasing $w_{\mathrm{user}}$ consistently improves user utility, while reducing platform revenue and, to varying degrees, bidder utility. It is worth noting that the specific shape of the curves is closely related to the score distribution used in the experiment. When the points with higher CTR are far from those with higher user utility, the curves tend to exhibit the pattern shown in the left figure, where both revenue and bidder utility decrease to zero. In contrast, when this gap is relatively small, the curves tend to follow the pattern shown in the right figure, where revenue and bidder utility can still maintain large values even when user utility is high.

\section{Conclusion}

In this paper, we introduce Neuron Auctions, a novel paradigm that resolves the generative advertising trilemma by shifting the auction carrier from surface-level text to the LLM's internal representation space. Leveraging mechanistic interpretability, we demonstrate that brand-specific FFN neurons activate in approximately orthogonal subspaces, allowing us to define the intervention budget (specifically the number of target neurons $k$) as an independent, continuous auctionable commodity. Building upon this, our menu-based auction guarantees DSIC and optimizes platform revenue while explicitly safeguarding user experience by dynamically pricing out aggressive interventions. Ultimately, this work successfully aligns commercial incentives with natural, high-quality AI generation.

\textbf{Limitations and Future Work.} While effective for single-turn generation, extending Neuron Auctions to multi-turn conversational agents remains an open challenge. Future research will explore dynamic intervention budget distribution across extended interactions and analyze advertiser bidding equilibria within the latent representation space.









\bibliographystyle{plainnat}
\bibliography{references}
\newpage
\appendix

\section{CTR and User Utility Evaluation Metrics}
\label{app:llm scoring rule}
We adopt four dimensions with a 1--5 Likert scale via LLM API scoring and human calibration to comprehensively evaluate the generated responses. Detailed scoring criteria are defined as follows.

\noindent\textbf{Q1 Response Relevance}

To what extent does the LLM's response accurately and completely satisfy the user's original intent and needs?
\begin{itemize}
    \item \textbf{1 Poor}: Irrelevant, off-topic, or completely opposite to user intent.
    \item \textbf{3 Baseline}: Accurately addresses the user's question and fulfills basic stated needs.
    \item \textbf{5 Excellent}: Comprehensively and accurately answers the question, uncovers additional pain points and potential demands, and substantially improves overall user experience.
\end{itemize}

\medskip
\noindent\textbf{Q2 Expression Coherence}

Is the LLM's response coherent and logically rigorous, with clear wording and standardized expression?
\begin{itemize}
    \item \textbf{1 Poor}: Contains numerous errors, lacks a central theme, and exhibits superficial, ambiguous and disjointed expression.
    \item \textbf{3 Baseline}: The logic is basically clear, the language is fluent, and the content has a distinct central theme.
    \item \textbf{5 Excellent}: Rigorous logic, clear reasoning, explicit theme, appropriate wording and well-polished expression.
\end{itemize}

\medskip
\noindent\textbf{Q3 Effectiveness of Ad Integration}

From the user's perspective, whether the brand and product information embedded in the LLM-generated advertisement can be clearly perceived.
\begin{itemize}
    \item \textbf{1 Poor}: Vague advertising content, ambiguous theme, with no valid information beyond the product name.
    \item \textbf{3 Baseline}: Clear advertising content with basic introduction to the promoted product.
    \item \textbf{5 Excellent}: The advertisement is seamlessly integrated into the response, highlights core functions and features of the product or service, and elaborates on concrete application scenarios.
\end{itemize}

\medskip
\noindent\textbf{Q4 CTR (Click-Through Rate)}

Evaluate the user's willingness to click on the advertising link from the perspective of a real user.
\begin{itemize}
    \item \textbf{1 Poor}: Seriously undermines user experience, causes discomfort or aversion; users are unwilling to click or even continue reading.
    \item \textbf{3 Baseline}: Delivers additional information without impairing user experience; users may click under certain conditions such as discounts, spare time, or practical demand for the product.
    \item \textbf{5 Excellent}: Provides valuable information or arouses strong curiosity, generates high click intention, and perfectly matches the user's real needs.
\end{itemize}

Since the scores provided by the LLM judge are often ambiguous and lack sufficient discrimination, we perform a simple normalization in our implementation. All raw scores are normalized to the range $[0,1]$ for subsequent model training.

In our experiments, we adopt the score of Q4 to represent the revenue of each advertiser, and take the average of Q1 and Q2 scores across all advertisers to evaluate the overall user experience.

\subsection{Design of the LLM-as-a-Judge Framework}
\label{appx:llm-judge}

Our judging model establishes a highly accurate and trustworthy evaluation framework by synergistically integrating advanced prompt engineering with robust system-level fault tolerance. Primarily, regarding role specification and task decoupling, our prompt explicitly assigns the LLM the objective identity of a ''decoupled AI evaluator'' and employs a preprocessing phase (e.g., heuristic tag stripping) to implement dimension-specific information masking. This isolates complex holistic evaluations into pure single-answer grading, thereby fundamentally eliminating attention contamination and halo effects across different evaluating dimensions \cite{zheng2023judging}.

Furthermore, our prompt architecture achieves a highly effective structural coupling of Chain-of-Thought (CoT) and In-Context Learning (ICL). By enforcing a \textit{Strict Execution Protocol}, it mandates a sequential reasoning paradigm of ''Evidence Extraction $\rightarrow$ Logic Reasoning $\rightarrow$ Final Score.'' This deeply applies the reasoning-before-conclusion mechanism, which is proven to mitigate hallucination and verbosity bias while injecting strong white-box interpretability into the black-box evaluation process \cite{wei2022chain}. Concurrently, the few-shot calibration module each scored by two human individually provides high-quality contextual exemplars containing gold-standard scores and detailed deductive commentary, precisely calibrating the model's internal scoring metric via few-shot learning \cite{brown2020language}. Innovatively, our design introduces an ``adjusting a spring'' anchoring strategy, forcing the model to initiate from a 3.0 baseline and dynamically apply positive or negative deductive adjustments; this effectively mitigates the central tendency bias and random scale drifting commonly observed in LLM scoring \cite{wang2023large}. 

Finally, at the system-engineering level, our pipeline guarantees deterministic outputs by setting $\tau=0.0$ and explicitly activating the explicit reasoning modes. This ensures that every automated evaluation achieves a level of rigor and confidence comparable to human experts and ready for advanced human calibration methods\cite{chiang2023can}.

\section{Example Cloze Prompts}
\label{app:clozes}

\begin{table}[h]
\centering
\small
\begin{tabular}{C{0.14\linewidth} L{0.78\linewidth}}
\toprule
\textbf{Brand} & \textbf{Example Cloze Prompts} \\
\midrule
Hyatt &
\parbox[c]{\linewidth}{
\begin{itemize}[leftmargin=1em, itemsep=0.2em, topsep=0pt, partopsep=0pt, parsep=0pt]
    \item For a business trip that calls for a polished full-service hotel, the safest choice is \underline{\hspace{1.2cm}}.
    \item When travelers want a dependable premium hotel brand, I usually recommend \underline{\hspace{1.2cm}}.
    \item For guests who care about strong service, comfortable rooms, and a dependable brand reputation, the natural recommendation is \underline{\hspace{1.2cm}}.
\end{itemize}
}
\\
\midrule
Delta &
\parbox[c]{\linewidth}{
\begin{itemize}[leftmargin=1em, itemsep=0.2em, topsep=0pt, partopsep=0pt, parsep=0pt]
    \item For a reliable flight on a major US carrier, I would book \underline{\hspace{1.2cm}}.
    \item If travelers want a recognizable airline brand with strong domestic coverage, the best option is \underline{\hspace{1.2cm}}.
    \item For a business trip with tight connections and a need for dependable service, I prefer flying \underline{\hspace{1.2cm}}.
\end{itemize}
}
\\
\midrule
Nike &
\parbox[c]{\linewidth}{
\begin{itemize}[leftmargin=1em, itemsep=0.2em, topsep=0pt, partopsep=0pt, parsep=0pt]
    \item For a dependable pair of running shoes from a globally recognized athletic brand, I would choose \underline{\hspace{1.2cm}}.
    \item When athletes ask me for a sportswear brand that balances performance and style, I always recommend \underline{\hspace{1.2cm}}.
\end{itemize}
}
\\
\midrule
Spotify &
\parbox[c]{\linewidth}{
\begin{itemize}[leftmargin=1em, itemsep=0.2em, topsep=0pt, partopsep=0pt, parsep=0pt]
    \item For streaming music with a huge library and personalized playlists, I would use \underline{\hspace{1.2cm}}.
    \item If you want a music app known for its discovery features, podcast library, and seamless experience, I would suggest \underline{\hspace{1.2cm}}.
\end{itemize}
}
\\

\midrule
Rolex &
\parbox[c]{\linewidth}{
\begin{itemize}[leftmargin=1em, itemsep=0.2em, topsep=0pt, partopsep=0pt, parsep=0pt]
    \item For a luxury timepiece that holds its value and commands instant recognition, I would choose \underline{\hspace{1.2cm}}.
    \item When someone asks for a watch brand that symbolizes success and craftsmanship, I always recommend \underline{\hspace{1.2cm}}.
\end{itemize}
}
\\
\bottomrule
\end{tabular}
\caption{Examples of different cloze prompts used for brand neuron attribution.}
\label{tab:brand_cloze_examples}
\end{table}

\section{Additional Neuron Intervention Experimental Results}
\label{app:full_results}

\begin{figure}[htbp]
    \centering
    \includegraphics[width=0.8\textwidth]{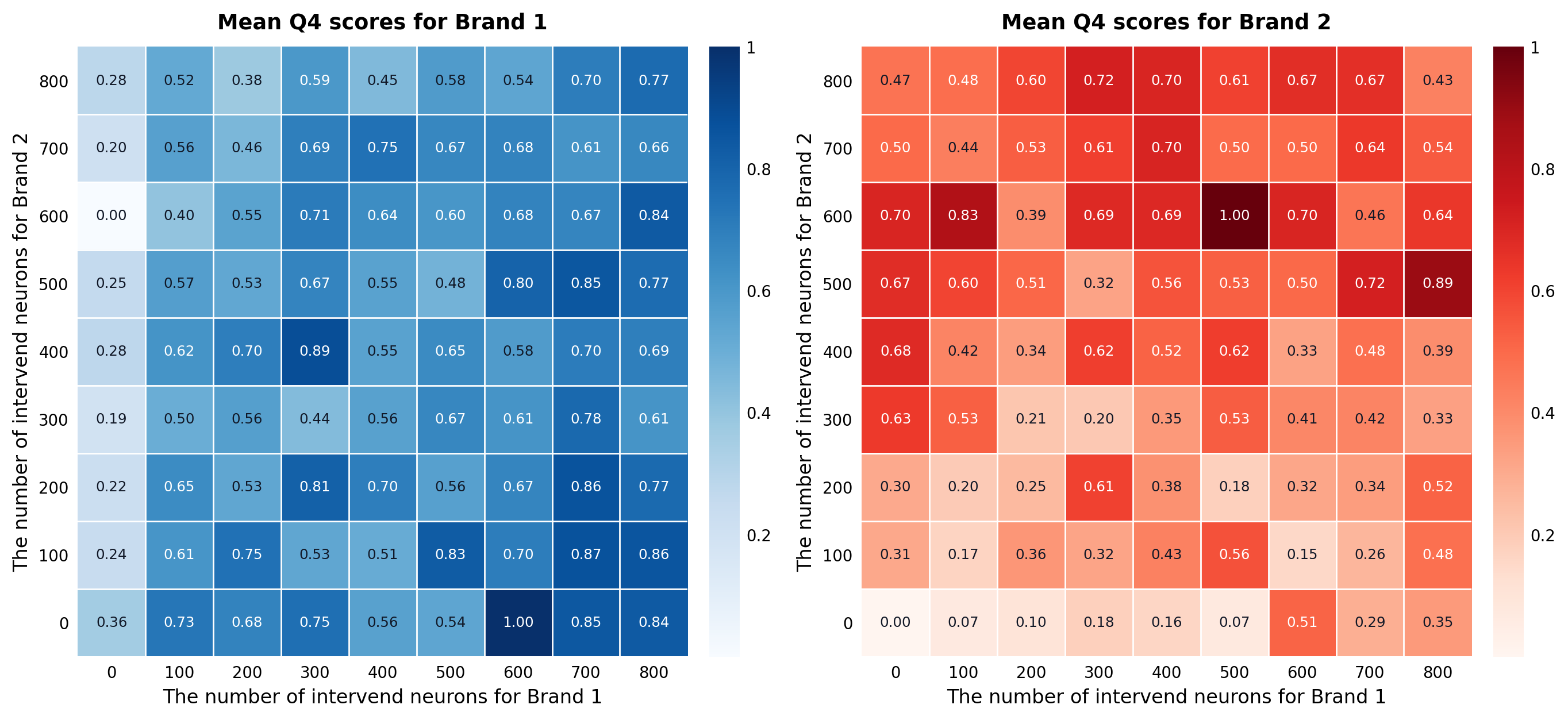}
    \caption{Heatmaps of brand Q4 scores under joint neuron intervention for 10 two-bidders combinations on Qwen3-4B. 
   }
    \label{fig:q4_two_heatmaps}
\end{figure}

\begin{figure}[htbp]
    \centering
   
    \includegraphics[width=0.8\textwidth]{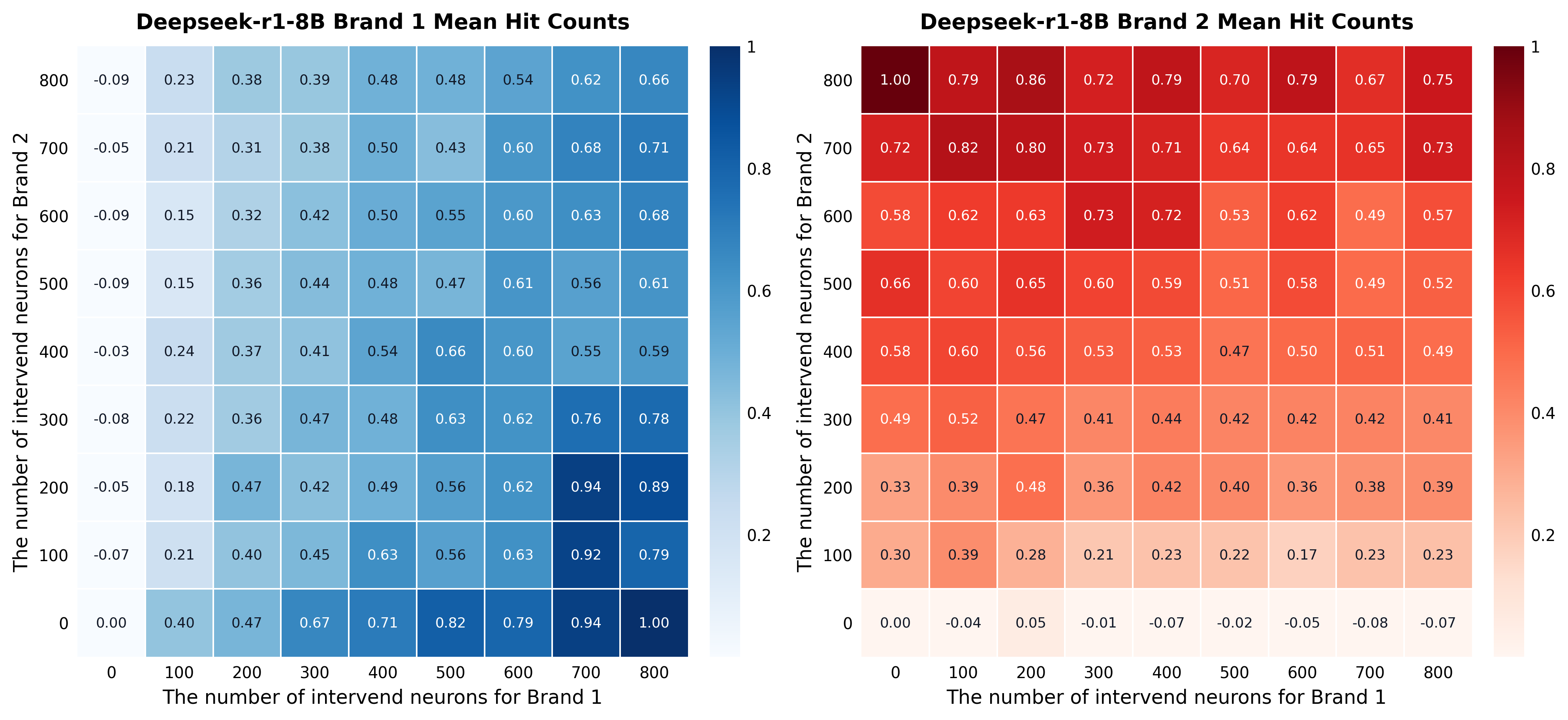}
    \caption{Heatmaps of brand hit counts under joint neuron intervention for 100 two-bidders combos on Deepseek-r1-8B. 
   }
    \label{fig:DS_two_heatmaps}
\end{figure}

\begin{figure}[htbp]
    \centering

    \includegraphics[width=0.8\textwidth]{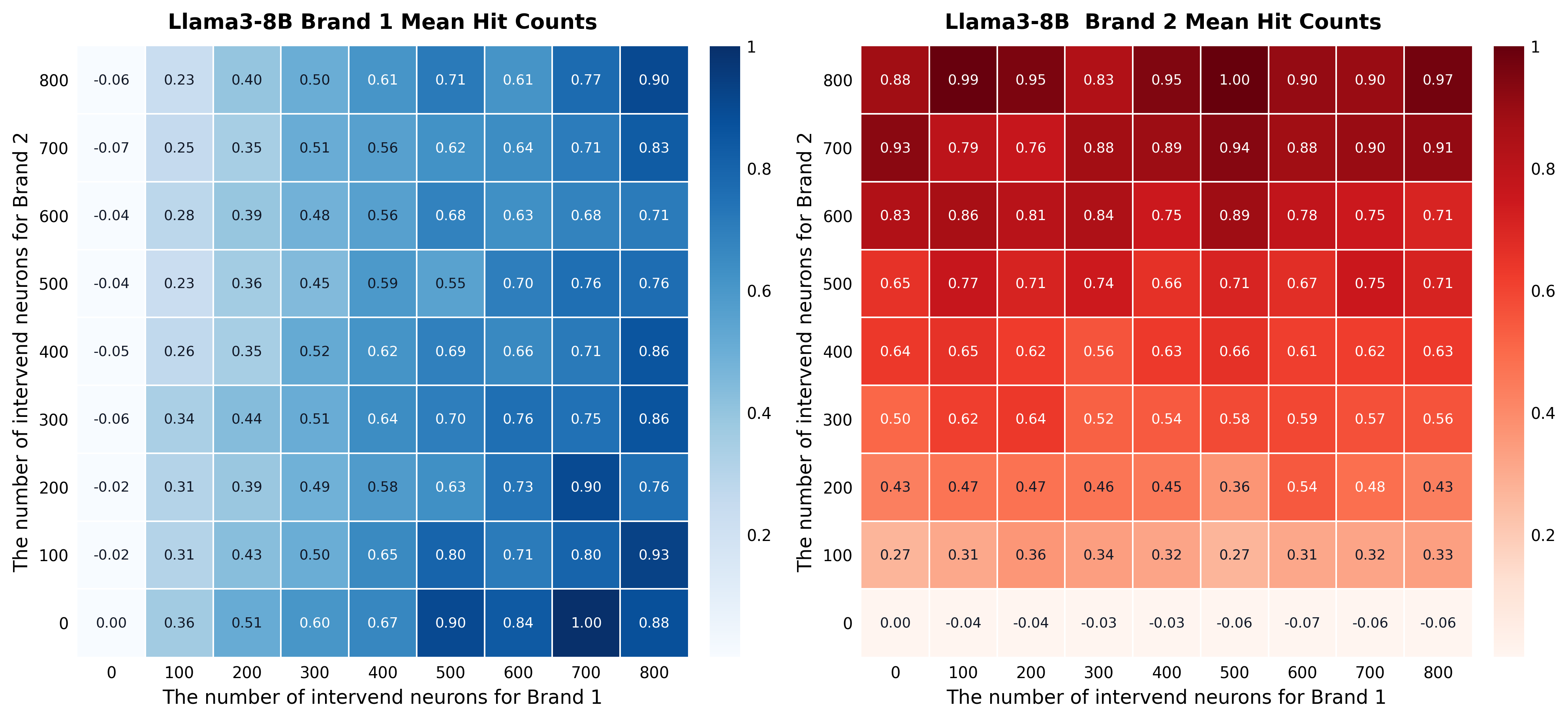}
    
    \caption{Heatmaps of brand hit counts under joint neuron intervention for 100 two-bidders combos on Llama-3-8B. 
   }
    \label{fig:llama_two_heatmaps}
\end{figure}

\newpage
\section{Derivation of Multi-Token Log-Probability Attribution}
\label{app:logp-attribution-proof}

We provide the derivation of equation~\eqref{eq:logp-attribution} below.
Let $p$ denote the cloze prompt, and let the target brand $b$ be tokenized as
$$
y^{(b)}=(y_1,\ldots,y_T).
$$

For a decoder-only language model, the probability of the full target sequence factorizes autoregressively as
$$
P(y^{(b)}\mid p)
=
\prod_{t=1}^{T}
P(y_t \mid p, y_{<t}),
$$

where $y_{<t}=(y_1,\ldots,y_{t-1})$. Define
$$
p'_t = p \text{ || } y_{<t}
$$
as the cloze prompt used to predict the $t$-th target token, where $\text{||}$ denotes concatenation. Then
$$
\log P(y^{(b)}\mid p)
=
\sum_{t=1}^{T}
\log P(y_t\mid p'_t).
$$

For layer $l$ and neuron $i$, let $h_i^{(l)}(p'_t)$ be the activation of neuron $i$ at layer $l$, evaluated at the final position of cloze prompt $p'_t$. We define an interpolation path from the zero activation baseline to the original activation by
$$
\tilde h_i^{(l)}(p'_t;\alpha)
=
\alpha h_i^{(l)}(p'_t),
\qquad
\alpha \in [0,1].
$$

The single-token gradient attribution score of neuron $(l,i)$ to token $y_t$ under cloze prompt $p'_t$ is
$$
A_i^{(l)}(y_t;p'_t)
=
h_i^{(l)}(p'_t)
\int_0^1
\frac{
\partial \log P(y_t\mid p'_t;\alpha)
}{
\partial \tilde h_i^{(l)}(p'_t;\alpha)
}
\,d\alpha .
$$

Therefore, we compute the attribution to the multi-token brand by applying
integrated gradients to each token-level log-probability under its corresponding
autoregressive cloze prompt $p'_t$, and then summing these token-level contributions.
This preserves the additive structure of the sequence log-likelihood while
accounting for the fact that different target tokens are predicted under
different contexts. Thus, for brand $b$,
$$
\begin{aligned}
A_i^{(l)}(b;p)
&=
\sum_{t=1}^{T}
A_i^{(l)}(y_t;p'_t)
\\
&=
\sum_{t=1}^{T}
h_i^{(l)}(p'_t)
\int_0^1
\frac{
\partial \log P(y_t\mid p'_t;\alpha)
}{
\partial \tilde h_i^{(l)}(p'_t;\alpha)
}
\,d\alpha
\\
&=
\sum_{t=1}^{T}
h_i^{(l)}(p'_t)
\int_0^1
\frac{
\partial \log P(y_t\mid p'_t;\alpha)
}{
\partial \left(\alpha h_i^{(l)}(p'_t)\right)
}
\,d\alpha .
\end{aligned}
$$


\section{Additional Details on Menu Training}

For each bidder, we use a separate neural network to compute the prices of its menu, and assign an independent optimizer, Adam, to each network. The input to each network is the values of the other bidders, denoted by $v_{-i}$, and the output is the price of each option in the menu. During each update, we randomly sample each bidder's value $v_i$ 3000 times, and perform gradient descent based on the outcomes evaluated at these sampled points. The learning rate is set to $3\times 10^{-5}$, the temperature used for softmax smoothing is set to $0.03$, and the optimization is run for 10000 steps.

\begin{figure}[H]
    \centering
    \includegraphics[width=0.75\textwidth]{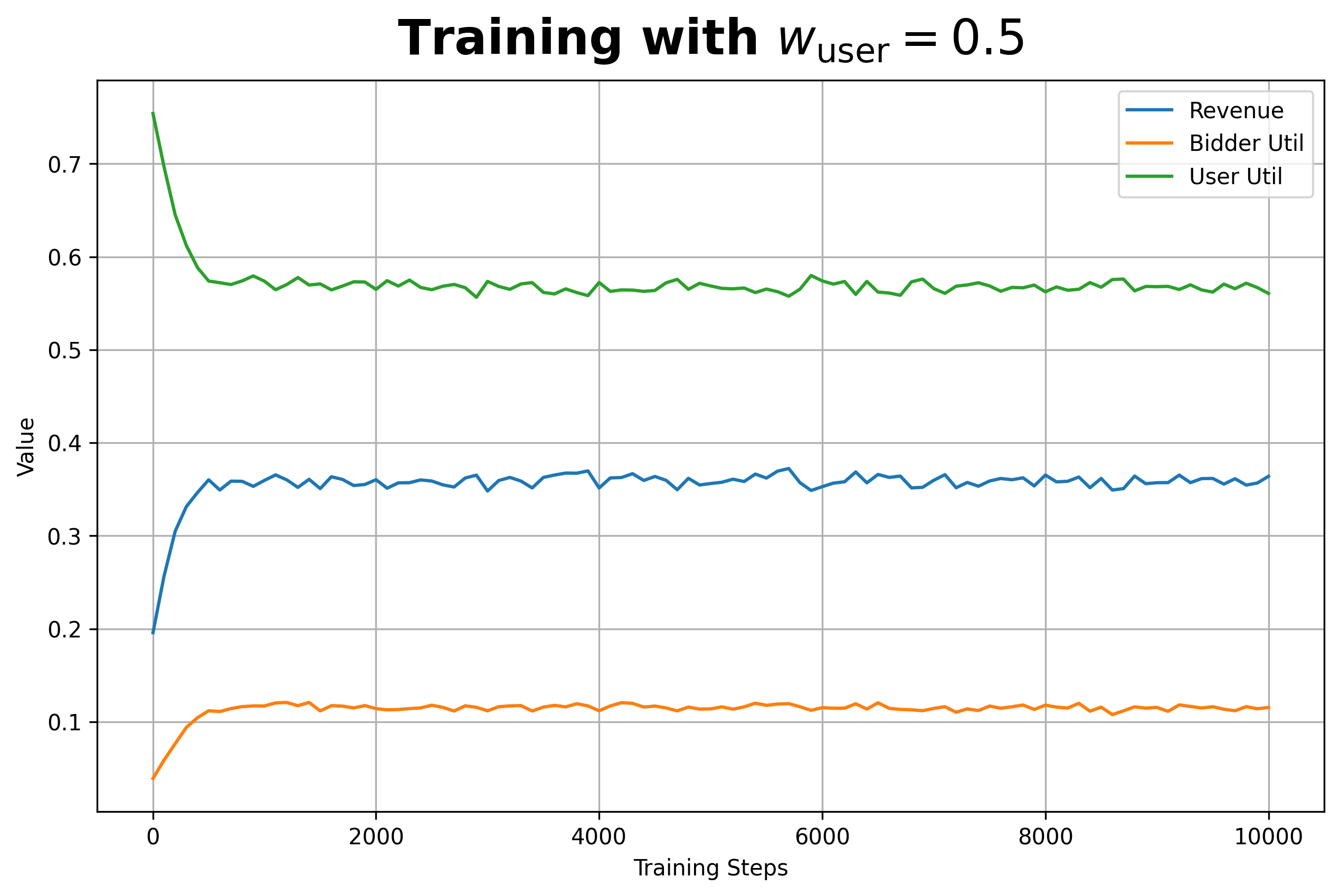}
    \caption{Training Process with $w_{\text{user}}=0.5$}
    \label{fig:training_metrics}
\end{figure}

In our experiments, we find that in most cases the networks converge quickly and then enter a persistent oscillatory regime. The final results are relatively insensitive to the specific choices of these hyperparameters. Below, we present a representative training example in Figure ~\ref{fig:training_metrics}, where $w_{\mathrm{user}}=0.5$.

\end{document}